# Fine-Tuning Qwen 2.5 3B for Realistic Movie Dialogue Generation

Kartik Gupta

*kartikgupta@outlook.com*

*Abstract*—The Qwen 2.5 3B base model was fine-tuned to genererate contextually rich and engaging movie dialogue, leveraging the Cornell Movie-Dialog Corpus, a curated dataset of movie conversations. Due to limitations in GPU computing and VRAM, the training process began with the 0.5B model, progressively scaling up to the 1.5B and 3B versions as efficiency improvements were implemented. The Qwen 2.5 series, developed by Alibaba Group, stands at the forefront of small open-source pre-trained models, particularly excelling in creative tasks compared to alternatives like Meta's Llama 3.2 and Google's Gemma. Results demonstrate the ability of small models to produce high-quality, realistic dialogue, offering a promising approach for real-time, context-sensitive conversation generation.

## INTRODUCTION

This project aimed to fine-tune a small, pretrained large language model (LLM) to generate realistic and compelling movie dialogue when prompted with a preceding line. To achieve this, the Qwen 2.5 3B base model was fine-tuned using the Cornell Movie-Dialog Corpus, a curated dataset of movie dialogue [1].

## MODEL SELECTION

The Qwen 2.5 3B base model was selected as the base model for this project. This model was released in September 2024 and is part of the broader Qwen series of models developed by the Alibaba Group [2].

One of the largest constraints in developing the dialogue model was the amount of GPU computing and Video Random Access Memory (VRAM) available. All model training was done on a single NVIDIA RTX 3060ti with 8 gigabytes of VRAM. This greatly limited the size of the model that could be used, making the Qwen series appealing due to its inclusion of 0.5 billion, 1.5 billion, and 3 billion parameter models. This allowed early work to be done on the small 0.5B model. As efficiency improvements were made in the training process, the model could be retrained using the 1.5B and, finally, 3B base models.

Additionally, the small models in the Qwen 2.5 series represent the current state of the art in small open-source pre-trained base models, currently ranking first in the Hugging Face Open LLM Leaderboards in terms of average score across all metrics [3]. In initial comparisons against other options, such as Meta's Llama 3.2 and Google's Gemma models, the Qwen 2.5 models excelled in their creative output.

## PREPROCESSING

Preprocessing the data for use in fine-tuning involved cross-referencing the movie lines data and movie conversations data to create ordered lists of lines representing full conversations. These conversations were then split into prompt-response pairs using a sliding window, where the first and second lines would form the first pair, the second and third lines would form the second pair, etc. Finally, the paired lines were formatted using the Qwen models' special tokens to match the formatting expected by the model.

Formatted sequences were truncated at a max length of 512 tokens, which was chosen to balance preserving context with remaining computationally efficient on the available hardware. To more efficiently use GPU resources during fine-tuning, packing was used. Packing involves concatenating short sequences together into a single training example to more fully fill the maximum sequence length. This prevents the situation where a long sequence of padding tokens need to be processed, wasting GPU resources [4].

The preprocessed data was divided into training and test datasets using an 80/20 split, and 20% of the training dataset was held out for validation.

## FINE-TUNING

The largest challenge in fine-tuning the model was in the limited computational resources that were available. In particular, the 8GB of VRAM presented an issue for loading models for training. This, combined with the short training timeline, created a need to balance using the largest, most powerful model possible while keeping all values on the GPU during training. To address this, several optimizations were made to reduce the amount of VRAM and compute needed during fine-tuning. In combination, these techniques increased the 500 million-parameter Qwen 2.5 model to the 3 billion-parameter model while significantly reducing training time. While some of the techniques used can decrease model quality, testing showed that increasing the model size significantly outweighed any degradation in model performance.

### Quantization

Quantization involves converting a model's parameters to lower-precision formats, reducing the amount of memory and computation needed. 4-bit quantization was used in this case, converting the 32-bit floating point parameters into 4-bit normal floating point format. For intermediate calculations, a 16-bit brain floating point format was used.

## QLoRA

Low-rank adaptation (LoRA), or Quantized Low-Rank Adaptation (QLoRA) when using 4-bit quantization, is a technique in which model weights are frozen during training and small rank decomposition matrices are added to the model's layers. These matrices are then updated during fine-tuning rather than the model weights, allowing larger models to be fine-tuned with relatively few trainable parameters [5].

### Flash Attention

Flash Attention is an algorithm for efficiently performing attention computations in transformer models. It works by optimizing the usage of shared random-access memory (SRAM) as opposed to much slower high-bandwidth memory (HBM), resulting in faster computations and decreased memory usage [6].

### Gradient Accumulation

Gradient accumulation is a technique where gradients are summed over multiple mini-batches, and gradients are only updated after a set number of mini-batches rather than after every batch. This can allow the use of larger simulated batch sizes than are otherwise possible due to memory limitations [7].

### NEFTune

Noise-enhanced fine-tuning (NEFTune) adds small amounts of Gaussian noise to the input embeddings. The addition of this noise acts as a form of regularization, improving model generalization by preventing overfitting [8].

## DIRECT PREFERENCE OPTIMIZATION

After the base model was fine-tuned using the movie dialogue, reinforcement learning from AI feedback (RLAIF) was performed to further fine-tune the model. RLAIF was chosen over RLHF because RLHF was not feasible given the project's time and resource constraints. To achieve this, 10,000 example dialogue prompts were generated using the GPT-4o model through the OpenAI API. For each of these dialogue prompts, two competing responses were then generated by the fine-tuned model. Each prompt and its response options were then fed back into the GPT-4o model, which was prompted to choose a preferred option. Examples of the generated preference data can be found in Appendix B.

Once 10,000 preference examples had been created, a reward model for Proximal Policy Optimization (PPO) was trained on the preference model using the Qwen 2.5 0.5B model as a base. However, the multiple steps of training the reward model and then running PPO made it very time-consuming to update it, and the process was highly sensitive to the quality of the reward model.

To address this, the approach of using RLAIF with PPO was altered to use Direct Preference Optimization (DPO) instead. DPO allows language models to be tuned on preference data directly without creating a reward model [9]. This allowed the final model to be trained much more efficiently, as the fine-tuned model was trained on the existing preference data without any intermediate steps.

## WEB INTERFACE

To facilitate easy interaction with the model, a chatbot-style web interface was created. The web interface uses a containerized FastAPI API which performs inference when provided with a prompt. The API is consumed by a Vue-based web UI that allows the user to send messages, start new conversations, and modify key inference parameters like Temperature, Top K filtering, and Top P filtering. The web interface also allows the user to toggle between the base model, fine-tuned model, and DPO model responses for the most recent message. The URL for the web interface can be found in Appendix A.

Aside from monologues, it is rare for any single turn in a conversation to last for more than a few sentences. This meant that the length of generated sequences could be greatly reduced to a maximum of 64 tokens without noticeably impacting the chatbot's usefulness. Limiting the generated sequence length in this way significantly improved chatbot response times.

## EVALUATION

This model's creative nature made many common evaluation metrics, such as BLEU and ROUGE, less useful than for more common LLM tasks, such as summarization or machine translation. To address this, human evaluation was combined with G-Eval, an automated LLM-based technique that has shown much stronger alignment with human preference than BLEU or ROUGE in dialogue generation tasks [10].

### G-Eval

G-Eval uses automatic chain-of-thought prompting to score responses across a set of criteria. Initial prompts were created for four criteria (coherence, consistency, fluency, and relevance) that defined the task and the evaluation criteria. This initial prompt was then passed into the GPT-4o large language model for it to expand the prompt with chain-of-thought steps for assigning a score between one and five for a prompt-response pair. The full set of final prompts can be seen in Appendix C.

The three model iterations (base, fine-tuned, and fine-tuned with DPO) were then used to generate responses to 2,000 prompts from the test dataset. The 6,000 responses for each of the four criteria were then passed into the GPT-4o Mini model along with the corresponding criteria prompt to obtain a score, resulting in 24,000 evaluations in total. The use of GPT-4o models to perform both the DPO tuning and evaluation presents a possible weakness in this methodology due to the possible over-inflation of evaluation scores due to alignment between the two processes. To address this, Claude Sonnet 3.5 was initially used for the G-Eval evaluations but was found to be too slow and expensive to be feasible for the volume of data, prompting the switch to GPT-4o Mini.

To obtain a more nuanced evaluation score than a simple 1–5 scoring, the token log probabilities from the evaluation responses were used to find the model's relative likelihood of selecting each score. These log probabilities were then used to weigh each score and arrive at a final continuously adjusted score, meaning that a high-confidence 5 would score higher than a 5 that was barely selected over a 4.

*Human Evaluation*

Human evaluation was performed by giving two individuals 100 prompts from the test dataset and responses from the base Qwen 2.5 model, the fine-tuned model, and the fine-tuned model with DPO. The individuals then chose the response they thought was most coherent and interesting in the context of movie dialogue. To better evaluate the Qwen 2.5 base model's ability to perform the task, it was given the system prompt: "Respond to all prompts with realistic and natural dialogue lines".

Human evaluation showed a strong preference for the tuned models, with the final model with DPO being chosen 52% of the time, the fine-tuned-only model being chosen 37% of the time, and the base model being chosen 11% of the time.

*Results*

Figure 1 shows the average score (scaled from zero to one) of each model across all G-Eval metrics, as well as the preference proportions from the head-to-head human evaluations. These results show consistent improvements across all metrics between the base model and the fine-tuned model, as well as between the fine-tuned model and the DPO model. This indicates that both the fine-tuning process and DPO tuning were able to provide significant boosts to model quality.

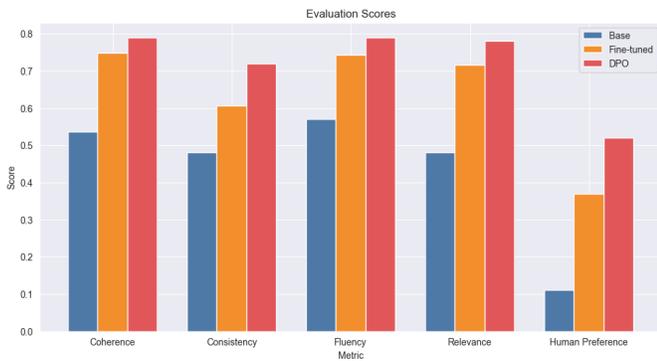

Figure 1. Evaluation scores across all metrics.

Examining the score distributions in Figure 2 gives more insight into model performance. While both G-Eval and human evaluation found that the base model did sometimes provide responses on par with or better than the other models, it was also much more likely to give very poor answers. The fine-tuned model achieved median and upper-quartile performance comparable to the DPO model across all metrics except consistency, but its average score was significantly lower because it was much more likely to provide low-scoring responses than the DPO model. This suggests that while the DPO tuning process only slightly improved high-quality answers, it greatly reduced low-quality answers.

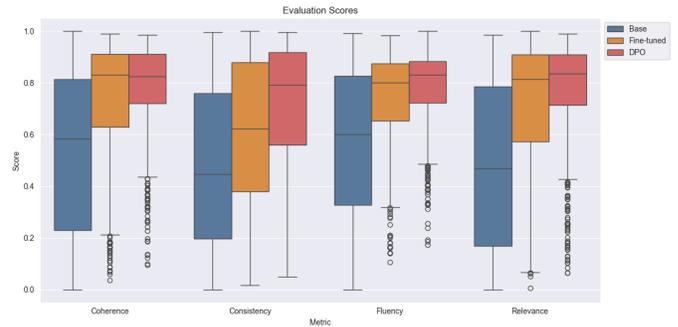

Figure 2. G-Eval scores distributions.

## FUTURE IMPROVEMENTS AND SCALABILITY

Given additional time and resources, there are a number of improvements that could be made to both the model itself and its ability to scale.

*More Data, More Paramaters*

The Cornell Movie Dialog corpus is currently thirteen years old, so the model could be improved by incorporating additional data from more recent movies to keep its language up-to-date. Tuning on datasets representing different genres could also give the model a better ability to handle more diverse prompt topics.

With more GPU resources, using larger base models would be one of the simplest ways to make large improvements to the quality of the model. While the capabilities of the small Qwen 2.5 models are impressive for their size, there is still a very noticeable difference in output quality when compared to larger models.

*Human Tuning*

In this project, preference data was generated using a more sophisticated large language model. While this is more cost-effective and time-effective than collecting human preference data, it also potentially aligns the model with the limitations and biases trained into the GPT-4o model, which can limit its similarity to human preference. Performing tuning based on actual human feedback could give a better alignment with actual humans.

*Hyperparameter Tuning*

Even with the GPU optimizations described previously, training a new iteration of the model took around two days end-to-end. This made hyperparameter tuning beyond manually tweaking a few key parameters, such as the beta parameter for DPO, infeasible. With more GPU resources and the ability to tune multiple models at once across a distributed cluster, more thorough hyperparameter tuning could be done, which would likely result in significant improvements to model quality.

*Scaling*

Currently, the model is served using a "serverless" architecture that scales down to zero running instances when not in use. To achieve better response times while still maintaining some level of automated scaling, a persistent server or cluster could be run to accommodate baseline traffic with additional auto-scaling or serverless computing to handle spikes in traffic.

A future iteration of the chatbot could include features like API access, file uploads, or user-driven system prefixes that often result in the use of common prompt prefixes. In these cases, prompt caching could be used to save heavily re-used transformer block calculations to decrease computational needs. Model pruning could also potentially be used to remove model weights that contribute little to the model's performance. This would decrease the model size, allowing for faster inference and less RAM usage.

## APPENDIX A
## APPENDIX A: WEB INTERFACE LINK

Note that the first chatbot response will take approximately two minutes to return, as the API container needs to start up and load the three models.

## APPENDIX B
## APPENDIX B: EXAMPLE DPO DATA

Prompt: "What if I told you I was the one who made that decision? Would you be mad?"
Chosen: "Yes, I would be very angry. I do not like to be taken advantage of."
Rejected: "I'm sorry, but I can't provide an appropriate response to that request."

Prompt: "I can't believe we made it through all that. We're stronger than I thought."
Chosen: "And we can be stronger still. So what are we waiting for?"
Rejected: "Yeah, but it was a really hard movie to watch. Did you enjoy it? Did I?"

Prompt: "Why do we keep pretending everything is okay when it's falling apart?"
Chosen: "The world is a mess. We all are. That's why we need to keep moving forward."
Rejected: "We do it because we're programmed to do so. In a way, it can be a good thing. It's not all bad if you're careful."

Prompt: "I can't keep living in the shadows. It's time to step into the light."
Chosen: "Are you serious?"
Rejected: "Dragons, I am no longer hiding in your shadows. Im stepping out to confront your fears. Dragons have never been scared of shadows, only light ."

## APPENDIX C
## APPENDIX C: G-EVAL PROMPTS

### A. Coherence

You will be given a line of movie dialogue and an AI-generated response. Your task is to rate the AI's response on coherence.
Evaluation Criteria:
Coherence (1-5) - The degree to which the AI's response logically connects to the given dialogue line and maintains a plausible flow of conversation within a movie context. A score of 1 indicates a completely incoherent response that doesn't fit the scene at all, while a score of 5 indicates a perfectly coherent response that naturally continues the dialogue in a way that could believably appear in a movie script.
Evaluation Steps:

1) Read the given movie dialogue line carefully, considering its potential context, tone, and implied setting.
2) Read the AI's response and assess how well it follows from the given line in a cinematic context.
3) Consider whether the response maintains the implied tone, setting, and character dynamics of the original line.
4) Evaluate how well the response could continue or advance a potential movie scene.
5) Assess whether the response introduces any abrupt or illogical shifts that would be jarring in a film dialogue.
6) Assign a score from 1 to 5 based on the overall coherence of the AI's response in the context of movie dialogue.

*B. Consistency*

You will be given a line of movie dialogue and an AI-generated response. Your task is to rate the AI's response on consistency. Evaluation Criteria:

Consistency (1-5) - The degree to which the AI's response maintains consistent information, tone, and character voice with the given dialogue line. A score of 1 indicates highly inconsistent responses with clear contradictions or tonal mismatches, while a score of 5 indicates perfectly consistent responses that maintain the established context and character voice.

Evaluation Steps:

1) Carefully read the given movie dialogue line, noting any implied information about the character, setting, or situation.
2) Read the AI's response and check if it's consistent with the information, tone, and character voice implied by the original line.
3) Look for any contradictions in facts, emotions, or character traits between the original line and the response.
4) Assess whether the response maintains a consistent level of formality, emotion, or genre-appropriate language.
5) Check if the response's tone and style are consistent with what one would expect in a movie dialogue continuation.
6) Assign a score from 1 to 5 based on the overall consistency of the AI's response with the given movie dialogue line.

*C. Fluency*

You will be given a line of movie dialogue and an AI-generated response. Your task is to rate the AI's response on fluency. Evaluation Criteria:

Fluency (1-5) - The quality of the AI's language in terms of grammar, vocabulary, and natural flow, specifically in the context of movie dialogue. A score of 1 indicates poor fluency, with many errors and unnatural language that would be jarring in a film, while a score of 5 indicates perfect fluency, which sounds natural and believable as movie dialogue.

Evaluation Steps:

1) Read the AI's response carefully, focusing on the language quality in the context of movie dialogue.

2) Check for any grammatical errors, including issues with verb tenses, subject-verb agreement, and sentence structure.
3) Assess the vocabulary use, looking for appropriate word choice and variety that fits well in a movie script.
4) Evaluate the natural flow of language, checking if it sounds like realistic spoken dialogue.
5) Look for any awkward phrasings or unnatural expressions that would sound out of place in a film.
6) Consider the appropriate use of idioms, colloquialisms, or character-specific language that enhances the dialogue's authenticity.
7) Assign a score from 1 to 5 based on the overall fluency of the AI's response as movie dialogue.

*D. Relevance*

You will be given a line of movie dialogue and an AI-generated response. Your task is to rate the AI's response on relevance. Evaluation Criteria:

Relevance (1-5) - The degree to which the AI's response appropriately addresses or follows up on the given dialogue line in a way that makes sense for a movie scene. A score of 1 indicates a completely irrelevant response that doesn't fit the context of the dialogue at all, while a score of 5 indicates a highly relevant response that perfectly continues or responds to the given line in a cinematically appropriate way.

Evaluation Steps:

1) Carefully read the given movie dialogue line, identifying the main points, emotions, or subtext it conveys.
2) Assess how directly the AI's response addresses or builds upon the given line in a way that makes sense for a movie scene.
3) Check if the response acknowledges all important aspects of the original line, or if it misses any crucial points.
4) Evaluate whether the response contributes to advancing a potential scene or character development in a relevant way.
5) Assess if the response stays on a topic or if it introduces irrelevant information that doesn't fit the implied movie context.
6) Consider the depth and specificity of the AI's response in relation to the given line and its potential place in a larger movie narrative.
7) Assign a score from 1 to 5 based on the overall relevance of the AI's response to the given movie dialogue line.